\documentclass[10pt,twocolumn,letterpaper]{article}

\usepackage{btas}
\usepackage{times}
\usepackage{epsfig}
\usepackage{graphicx}
\usepackage{amsmath}
\usepackage{amssymb}
\usepackage{gensymb}
\usepackage{subcaption}
\usepackage{url}

\usepackage{array}
\usepackage{multirow}
\usepackage{textcomp}
\usepackage{array}
\usepackage{colortbl}
\usepackage[table]{xcolor}
\usepackage{booktabs}

\definecolor{myblue}{rgb}{0,0,1}
\definecolor{myred}{rgb}{0.8, 0, 0}
\definecolor{mygreen}{rgb}{0, 0.6, 0}

\btasfinalcopy 

\ifbtasfinal\fi
\begin{document}

\title{Presentation Attack Detection for Cadaver Iris}

\author{Mateusz Trokielewicz\\
Institute of Control and Computation Engineering\\
Warsaw University of Technology\\
Nowowiejska 15/19, 00665 Warsaw, Poland\\
{\tt\small M.Trokielewicz@elka.pw.edu.pl}
\and
Adam Czajka\\
Department of Computer Science and Engineering\\
University of Notre Dame\\
IN, USA\\
{\tt\small aczajka@nd.edu}
\and
Piotr Maciejewicz\\
Department of Ophthalmology\\
Medical University of Warsaw\\
Lindleya 4, 02005 Warsaw, Poland\\
{\tt\small piotr.maciejewicz@wum.edu.pl}
}

\maketitle
\thispagestyle{empty}

\begin{abstract}
This paper presents a deep-learning-based method for iris presentation attack detection (PAD) when iris images are obtained from deceased people. Post-mortem iris recognition, despite being a potentially useful method that could aid forensic identification, can also pose challenges when used inappropriately, \ie utilizing a dead organ of a person in an unauthorized way. Our approach is based on the VGG-16 architecture fine-tuned with a database of 574 post-mortem, near-infrared iris images from the Warsaw-BioBase-PostMortem-Iris-v1 database, complemented by a dataset of 256 images of live irises, collected within the scope of this study. Experiments described in this paper show that our approach is able to correctly classify iris images as either representing a live or a dead eye in almost 99\% of the trials, averaged over 20 subject-disjoint, train/test splits. We also show that the post-mortem iris detection accuracy increases as time since death elapses, and that we are able to construct a classification system with APCER=0\%@BPCER$\approx$1\% (Attack Presentation and Bona Fide Presentation Classification Error Rates, respectively) when only post-mortem samples collected at least 16 hours post-mortem are considered. Since acquisitions of ante- and post-mortem samples differ significantly, we applied countermeasures to minimize bias in our classification methodology caused by image properties that are not related to the PAD. This included using the same iris sensor in collection of ante- and post-mortem samples, and analysis of class activation maps to ensure that discriminant iris regions utilized by our classifier are related to properties of the eye, and not to those of the acquisition protocol. This paper offers the first known to us PAD method in a post-mortem setting, together with an explanation of the decisions made by the convolutional neural network. Along with the paper we offer source codes, weights of the trained network, and a dataset of live iris images to facilitate reproducibility and further research.  \let\thefootnote\relax\footnote{978-1-5386-7180-1/18/\$31.00 \copyright2018 IEEE}
\end{abstract}

\section{Introduction}
Post-mortem biometric identification is a field of study well established in the scientific community, among forensics professionals, but also in popular culture, with severed thumbs and plucked eyeballs being depicted in the big screen disturbingly often. With increasing importance that biometric authentication gains in our daily lives, fears are increasingly common among users, regarding the possibility of unauthorized access to our data, identity, or assets after our demise. Law enforcement officers in the U.S. are reportedly already using the fingerprints of the deceased to unlock the suspects' iPhones \cite{ColdTouchID}, which immediately brings up the topic of whether liveness detection should be one of the components of Presentation Attack Detection implemented in such devices. With a constantly growing market share of iris recognition, and recent research proving that iris biometrics in a post-mortem scenario can be  viable \cite{TrokielewiczPostMortemICB2016,TrokielewiczPostMortemBTAS2016,BolmeBTAS2016,TrokielewiczArxiv2018}, these concerns are also becoming true for iris. In a recent interview for the {\it IEEE Spectrum} online magazine, Czajka discussed the issue of liveness detection, which is crucial in those cases when \emph{we don't want} our biometric traits to be used after death \cite{ColdIrisSPECTRUM}. 

To our best knowledge, there are no prior papers or published research regarding the topic of discerning live irises from dead ones. This work thus offers the first study of iris liveness detection in a post-mortem scenario and offers the following {\bf contributions to the state-of-the-art}:   

\begin{itemize}
	\item a method for iris liveness detection in a post-mortem setting, using a static iris image and based on a deep convolutional neural network (DCNN) fine-tuned with a dataset of post-mortem and live iris images,
	\item an analysis of iris regions or features being most used by the network when providing its decision, employing class activation mapping,
	\item a complementary dataset of live iris images, collected with the same equipment as the existing post-mortem iris images dataset,
	\item source codes and network weights for the offered solution.
\end{itemize} 

Source codes, network weights, and a complementary dataset of live iris images can be obtained at \verb|http://zbum.ia.pw.edu.pl/EN/node/46|.

This article is laid out as follows. Section \ref{sec:related} presents a short discussion of presentation attack detection methods for iris recognition, applications of deep learning for classification tasks, and techniques for explaining the reasoning of a DCNN-based classifier. Section \ref{sec:data} familiarizes the Reader with the dataset of live and post-mortem iris images used in this study. It also describes an initial experiment aiming at explaining the decisions that deep neural network is making when classifying iris images. Sections \ref{sec:experiments} and \ref{sec:results} provide an overview of the experimental methodology and results, respectively. Finally, relevant conclusions are given in Section \ref{sec:conclusions}.

\section{Related work}
\label{sec:related}

\subsection{Presentation Attack Detection in iris recognition} 

Presentation attack detection is already a well established area in the field of biometrics-related research. Existing methods include detection of fake representations of irises (paper printouts, textured contact lenses, prosthetic eyes, displays), or a non-conformant use of an actual eye. The most popular techniques used in iris PAD use various image texture descriptors (Binarized Statistical Image Features (BSIF) \cite{Komulainen_IJCB_2014}, Local Binary Patterns (LBP) \cite{Doyle_ICB_2013}, Binary Gabor Patterns (BGP) \cite{Lovish_CAIP_2015}, Local Contrast-Phase Descriptor (LCPD) \cite{Gragnaniello_TIFS_2015}, Local Phase Quantization (LPQ) \cite{Sequeira_TSP_2016}, Scale Invariant Descriptor (SID) \cite{Gragnaniello_SITIS_2014}, Scale Invariant Feature Transform (SIFT) and DAISY \cite{Pala_CVPR_2017}, Weber Local Descriptor (WLD) \cite{Gragnaniello_TIFS_2015}, or Wavelet Packet Transform (WPT) \cite{Chen_PRL_2012}), image quality descriptors \cite{Galbally_Handbook_2016}, or deep-learning-based techniques \cite{Menotti:TIFS:2015,He_BTAS_2016,Pala_CVPR_2017,Raghavendra_WACV_2017}. If hardware adaptations are possible one may consider multi-spectral analysis \cite{Thavalengal_TCE_2016} or estimation of three-dimensional iris features \cite{Pacut_ICCST_2006,Hughes_HICSS_2013} for PAD. Making the PAD more complex, one may consider measuring micro-movements of an eyeball, either using Eulerian video magnification \cite{Raja_TIFS_2015} or by using an eye-tracking device \cite{Rigas_PRL_2015}, or measuring  pupil dynamics \cite{Czajka_TIFS_2015}. An extensive review of the state of the art in PAD for iris recognition, including a systematization of attack methodologies and countermeasures proposed, can be found in \cite{CzajkaPADSurvey2018}, and independent evaluations of algorithms detecting iris paper printouts and textured contact lenses can be studied from the LivDet-Iris competition series (\url{http://livdet.org/}), which has had the last edition in 2017 \cite{Yambay2017}.

Despite the abundance of research and proposed methods, there are still no published papers that would explore the concept of liveness detection in a scenario when cadaver (post-mortem) eyes are used to perform a presentation attack on the biometric sensor. However, one can still envisage such situation, in which a dead eye is used with an unsupervised biometric system to gain an unauthorized access to the assets the system is protecting.

\subsection{Deep convolutional nets in image classification}

Deep convolutional neural networks (DCNN) are useful in solving selected groups of computer vision problems, such as image classification \cite{VGGSimonyanCNNsForRecognition2014}, semantic segmentation \cite{SemanticSegmentationREVIEWarxiv}, automatic image captioning \cite{JohnsonKarpathyDenseCapCVPR2016}, or visual question answering \cite{RenVQAExploringModelsNIPS2015}. 

These methods learn model parameters (network weights and biases) and hyper-parameters (network architecture details and training constraints) by guessing them from the data that is made available to the model during the training phase. Thus, the model learns the data itself, constituting a \emph{feature-learnt} or \emph{data-driven} approach. This is opposed to finding the model parameters by obtaining prior knowledge about the object being recognized and fine-tuning much fewer parameters of a less complicated model, which may be called \emph{feature engineering} or \emph{hand-crafted} approaches. Data-driven approaches, such as these involving neural networks, offer significant advantages over hand-crafted ones in situations where the knowledge about the subject is either limited or difficult to be put into relatively simple mathematical rules that would enable building a feature-engineered solver. 

This can be the case with processing post-mortem iris samples, either to recognize a person or to recognize a presentation attack. Trokielewicz \etal revealed that although post-mortem iris recognition is, to some extent, possible with current software, it also poses new challenges that we do not yet have solutions to \cite{TrokielewiczPostMortemBTAS2016}. Most importantly, there currently are no mathematical models that would explain the iris' behavior over the course of post-mortem time horizon, \ie quantify and predict the changes that the iris may undergo after one's demise. Therefore, when aiming at discerning live irises from dead ones, a potentially promising way of solving this problem is to rely on the feature-learnt approach that utilizes the existing datasets of post-mortem and live iris images to teach itself to give the correct answer.       

This has already proven promising by enabling us to propose a post-mortem iris image segmentation method based on a DCNN, which achieves performance superior to the typical iris recognition method \cite{TrokielewiczPostMortemSegmentationIWBF2018}. 

\subsection{Explaining deep networks' decisions}
\label{sec:CAM}

Despite spectacular successes in computer vision tasks, DCNNs all have a significant drawback, namely their inability to provide an intuitive, human-understandable explanation for their decisions. Hence, although the performance of these solutions may be excellent, in their basic designs they lack interpretability. This has at least two potential downsides: 1) if humans do not know how an expert system (\eg a self-driving car software) works, they will not trust it, and 2) if the creators do not know how the system works, they cannot improve it.    

Zhou \etal \cite{CAMZhou} propose a technique called {\it class activation mapping}, or CAM, for identification of discriminative image regions, \ie these, which are decisive when it comes to the classification output. This is said to be feasible despite training the network only with image-level labels, and not dense, pixel-wise labels. The authors achieve this by dropping fully-connected layers in the popular network architectures (ALexNet, GoogLeNet, VGG), and replacing them with global average pooling layers followed only by a fully-connected output softmax layer. To increase spatial resolution of the mappings, some convolutional layers are also removed from the architectures, resulting in $14\times14$ or $15\times15$ outputs from the convolutional part of the respective network. This approach enables highlighting image regions that are important for discrimination, and even localize regions responsible for detecting patterns, such as text, and even higher-order concepts. 

Selvaraju \etal introduce improvement over the Zhou's method with Grad-CAM \cite{GradCAMSelvaraju}, a technique similar to CAM, but not requiring any changes to the network's architecture, and thus easier to use on the already trained models. This approach produces coarse localization heat-maps highlighting the regions that are considered important by the network when predicting the concept in the image. Also, by combining these low-resolution maps with high-resolution visualization of features learned by the network, obtained from guided back-propagation introduced by Springenberg \etal in \cite{GuidedBackPropSpringenberg}, it is possible to obtain a more fine-grained importance maps, which apart from highlighting a coarse \emph{region} of the image that is considered discriminatory, also allows insight into which \emph{features} are important. 

These techniques can be important for two reasons, both of them being explored in our paper. First, class activation mapping can help analyze the potential bias in the raw data, that can interfere with the network training, causing the model to learn features that are not directly related to the task at hand. For instance, learning the presence of metal retractors used to open cadaver eyes, and missing in live eyes, ends up with perfect accuracy albeit with no relation to PAD accuracy. Second, we hope to gain some knowledge regarding the iris/eye features being employed by the network for discriminating between live and dead irises.

\section{Experimental dataset}
\label{sec:data}
\subsection{Post-mortem iris images}
For the purpose of this study, we used the only, known to us, publicly available Warsaw-BioBase-PostMortem-Iris-v1 dataset, which gathers 1,330 post-mortem iris images collected from 17 individuals during various times after death (from 5 hours up to 34 days) \cite{WarsawColdIris1}. These samples represent ocular regions of recently deceased subjects. In addition to typical, near-infrared (NIR) iris images collected from cadavers, high quality visible light images are also available, however, for the purpose of this study we only employed the former, as NIR samples are usually employed in commercial, deployed iris recognition systems. There are 574 NIR images available in the dataset, most of them captured up to two days after death, but some of the samples extend up to 814 hours after death.  

\subsection{Images of live irises}

Since post-mortem part of the dataset used in this study does not offer any ante-mortem samples, or reference images of live individuals, we had to collect a complementary dataset of iris images collected from live people. To mimic the original acquisition protocol as closely as possible, and thus to minimize the bias in training the DCNN, we have employed the same iris camera as was used in the post-mortem counterpart, namely the IriTech IriShield M2120U.

\begin{figure*}[t]
	\centering
	\includegraphics[width=0.32\textwidth]{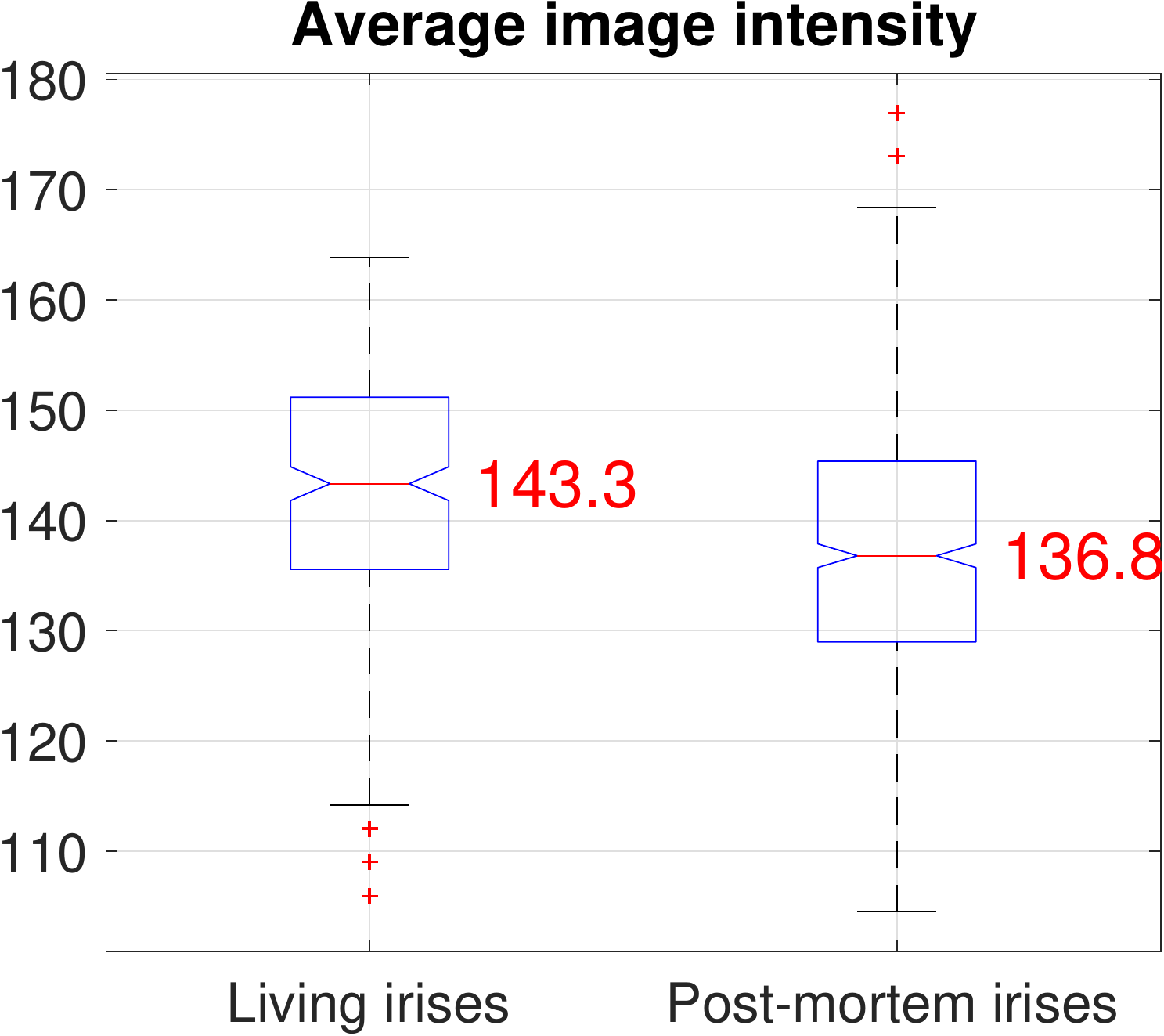}\hskip3mm
	\includegraphics[width=0.32\textwidth]{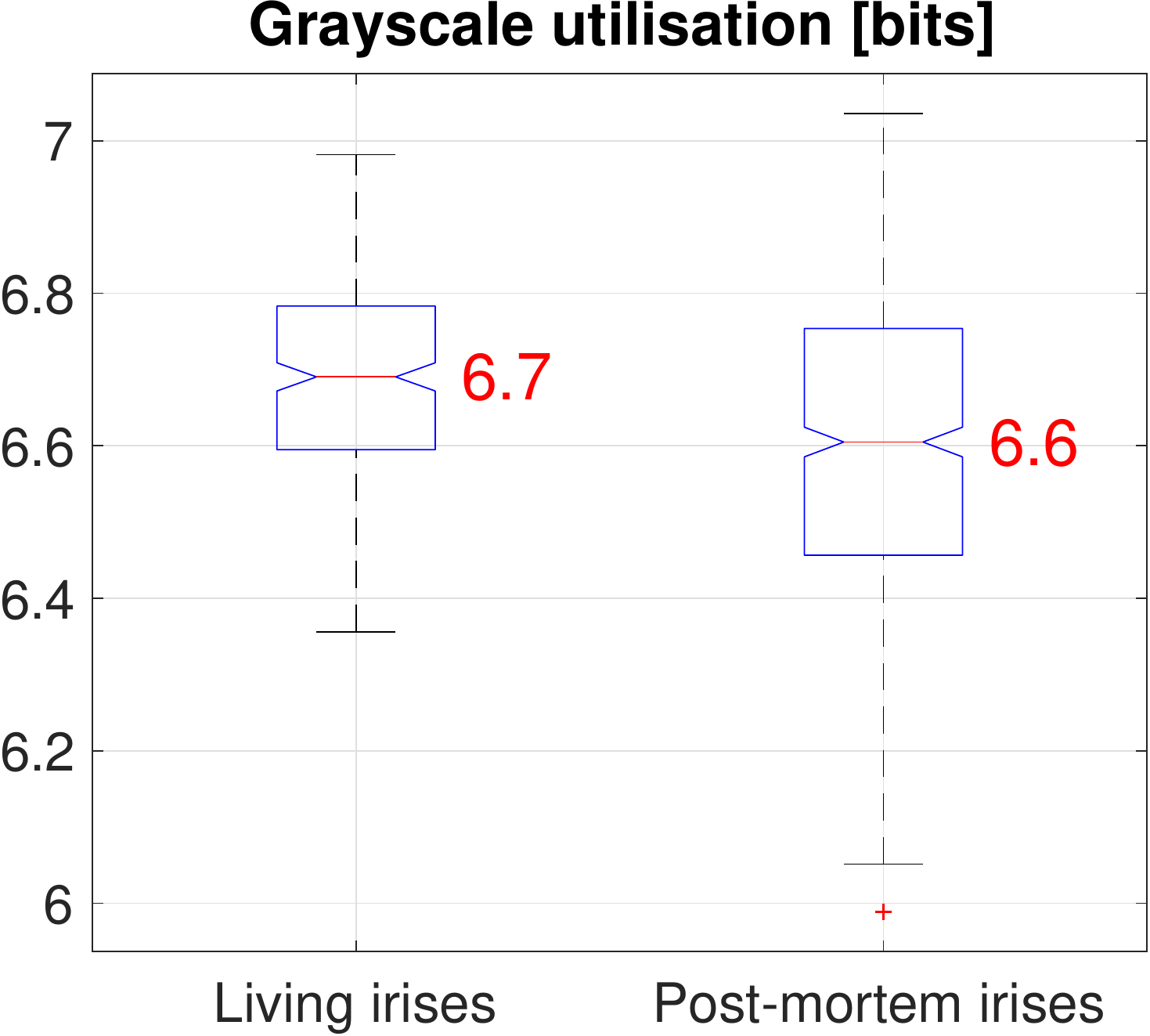}\hskip3mm
	\includegraphics[width=0.32\textwidth]{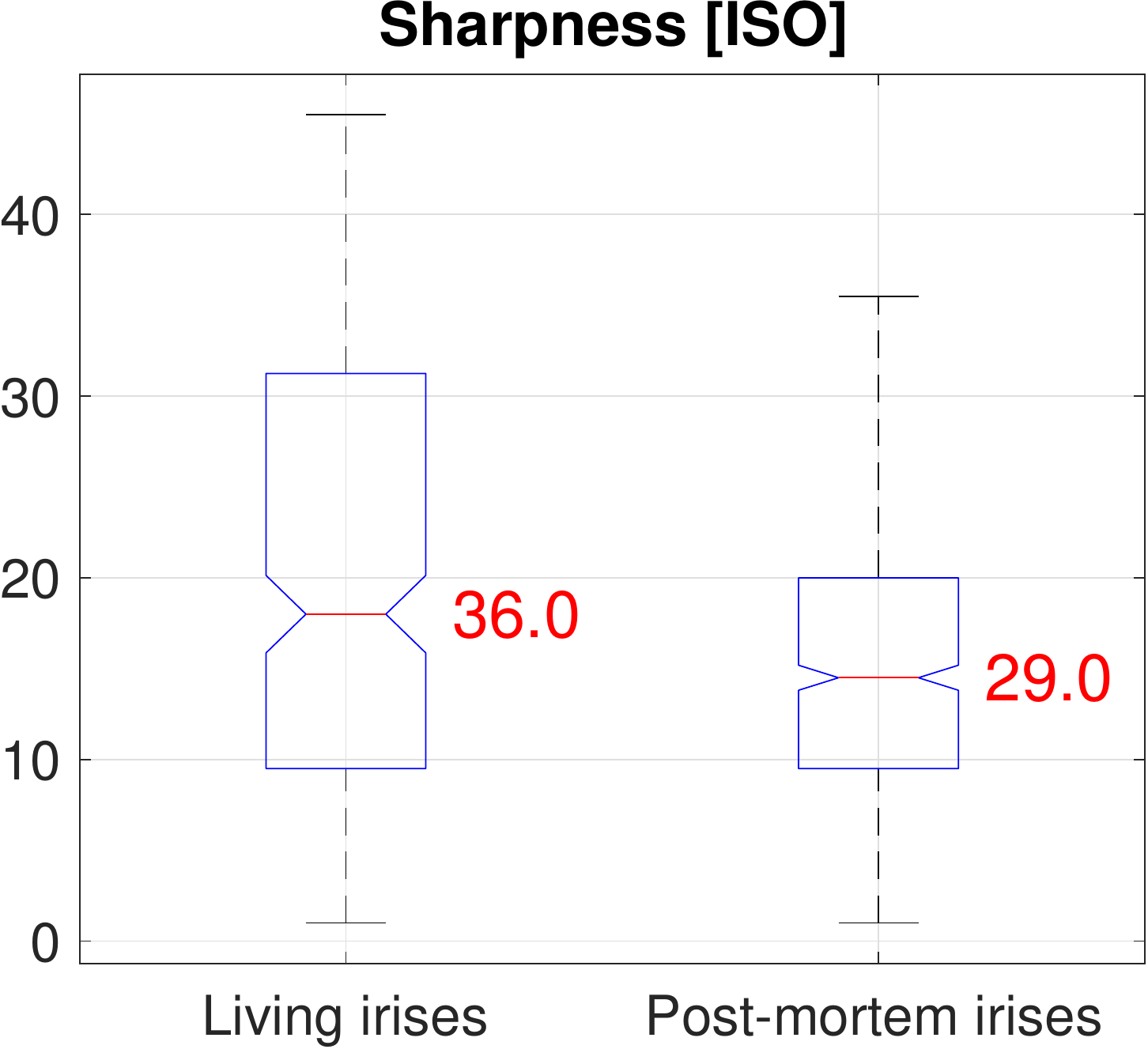}
	\caption{Boxplots representing differences in image quality metrics between the two datasets. Median values are shown in red, height of each boxes corresponds to an inter-quartile range (IQR) spanning from the first (Q1) to the third (Q3) quartile, whiskers span from Q1-1.5*IQR to Q3+1.5*IQR, and outliers are shown as crosses.}
	\label{fig:boxplots}
\end{figure*}

\begin{figure}[t!]
	\centering
	\includegraphics[width=0.235\textwidth]{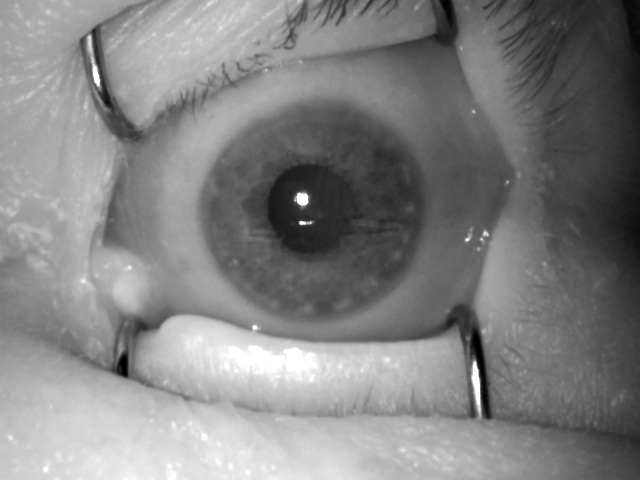}\hskip1mm
	\includegraphics[width=0.235\textwidth]{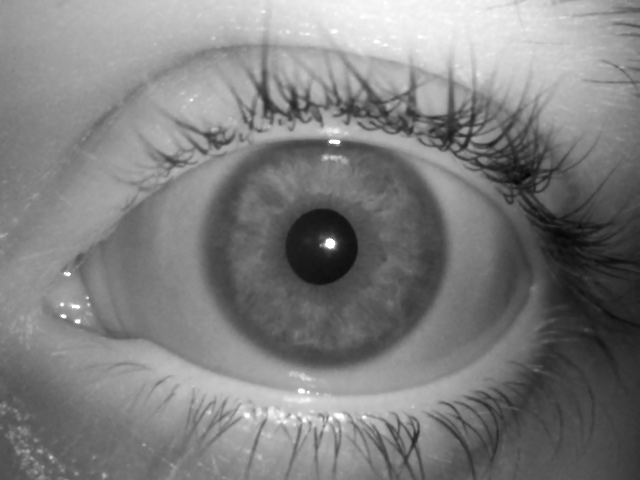}\\\vskip1mm
	\includegraphics[width=0.115\textwidth]{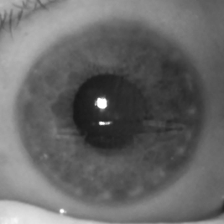}\hskip1mm
	\includegraphics[width=0.115\textwidth]{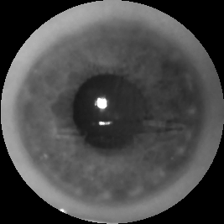}\hskip1mm
	\includegraphics[width=0.115\textwidth]{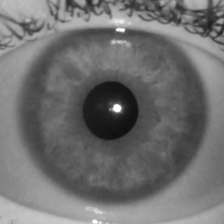}\hskip1mm
	\includegraphics[width=0.115\textwidth]{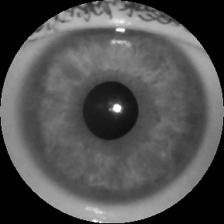}
	\caption{Example iris images obtained from a dead \textbf{(left)} and a live subject \textbf{(right)}: original images \textbf{(top row)} and their cropped and cropped\_masked versions \textbf{(bottom row)}.}
	\label{fig:samplesFromDataset}
\end{figure}

\subsection{Analysis of potential data bias}

We are aware that even despite our best efforts to keep the acquisition protocols as alike as possible, there will be some bias in the data. The aim of this Section is thus to carefully examine these differences, discuss their importance and impact on the experiments, and to propose countermeasures, where possible. To quantify the variations between images coming from the two datasets that may originate in difference between camera operators or the environment, we have performed the calculations of three covariates related to iris quality, namely: average intensity, grayscale utilization (image histogram entropy), and image sharpness, the latter two being suggested by the ISO/IEC standard on iris image quality \cite{ISO2}. These covariates are defined as follows:

\begin{itemize}
	\item \textbf{average image intensity}: 
	
	$$AI=\frac{1}{N}\sum_{i=1}^{n}\sum_{j=1}^{m}I_{ij}$$ 
	
	where $I_{ij}$ is the pixel intensity and $n,m$ is the image size in pixels;
	
	\item \textbf{grayscale utilization}, or the entropy of the iris image histogram $H$ measured in bits, examines pixel values of an iris image to calculate a spread of intensity values and assess whether the image is properly exposed:
	
	 $$H=-\sum_{i=1}^{256}p_ilog_2p_i$$
	
	 where $p_i$ is the probability of each gray level $i$ occurring in the image, hence, the total count of pixels at gray level $i$, divided by the total number of pixels in the image \cite{ISO2};

	\item \textbf{sharpness}, determined by the power resulting from filtering the image with a Laplacian of Gaussian kernel; for brevity, we do not reproduce the formulae here, and instead refer the Reader to the original ISO/IEC documentation \cite{ISO2}.
\end{itemize}

Results of these calculations are shown in Fig. \ref{fig:boxplots}. Notably, only the \emph{sharpness} covariate differs largely between the two datasets, as post-mortem iris images have lower sharpness on average. This can be a result of a combination of factors, such as: more difficult collection environment (\eg a hospital mortuary), a less experienced operator (\eg medical staff), limited time, and such. For completeness, to provide formal statistical analysis, we ran a Wilcoxon rank-sum test for each pair of covariates, which revealed that there are statistically significant differences between the subsets of live and post-mortem iris images, as the null hypothesis stating that the compared scores are samples from continuous distributions with equal median was rejected at significance level $\alpha=0.05$ in all three cases. However, all three covariates do not provide enough differentiation between the images coming from different datasets, to themselves serve as features for presentation attack detection detection.

\subsection{Assessing data bias with Grad-CAM}

Despite our efforts to follow the same collection guidelines as those of \cite{WarsawColdIris1}, there are still some differences between the respective samples that originate in different presentation of the biometric itself in live and post-mortem scenarios. This is most notably related to the appearance of eyelids, which in the post-mortem data are often pulled apart with a metal retracting device to keep the eye open for image acquisition. To at least partially mitigate these differences, the subjects participating in the collection of the reference data were asked to open their eyes as widely as possible. However, the presence of metal parts of the medical equipment is still an issue, as these appear in post-mortem cases and do not appear in live cases.

To examine whether these metal retractors will serve as cues for the DCNN when it is trained to discern post-mortem irises from the live ones, we have employed the class activation mapping technique described in Section \ref{sec:CAM} by performing an additional experiment, in which the network, same as described later in Sec. \ref{sec:experiments}, was trained with the original, uncropped images, as shown in Fig. \ref{fig:samplesFromDataset}, top row. This initial implementation of the network was done in Keras \cite{Keras}, using an adapted code from \cite{GradCAMCode}. 

\begin{figure}[t!]
	\centering
	\includegraphics[width=0.115\textwidth]{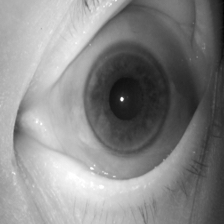}\hskip1mm
	\includegraphics[width=0.115\textwidth]{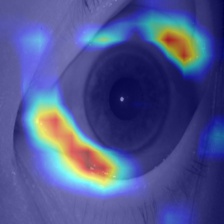}\hskip1mm
	\includegraphics[width=0.115\textwidth]{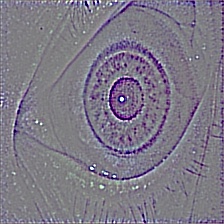}\hskip1mm
	\includegraphics[width=0.115\textwidth]{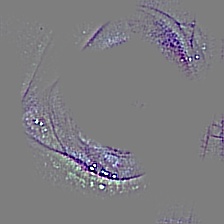}\\\vskip1mm
	\includegraphics[width=0.115\textwidth]{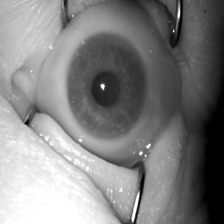}\hskip1mm
	\includegraphics[width=0.115\textwidth]{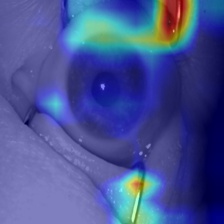}\hskip1mm
	\includegraphics[width=0.115\textwidth]{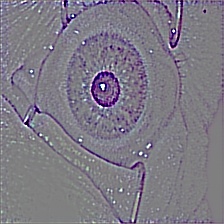}\hskip1mm
	\includegraphics[width=0.115\textwidth]{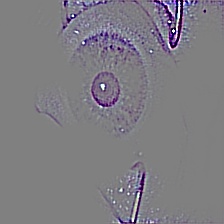}
	\caption{Example class activation maps obtained using the Grad-CAM technique for samples from the original, unmodified dataset, with a model trained on the original dataset. From left to right: (1) original image, (2) Grad-CAM, (3) guided back-propagation, (4) a combination of (2) and (3). \textbf{Post-mortem samples are represented here, both correctly classified}.} 
	\label{fig:bias-analysis-cam-cold}
\end{figure}

\begin{figure}[t!]
	\centering
	\includegraphics[width=0.115\textwidth]{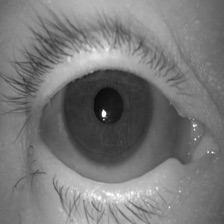}\hskip1mm
	\includegraphics[width=0.115\textwidth]{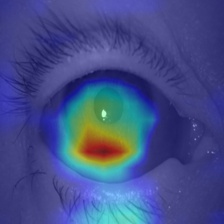}\hskip1mm
	\includegraphics[width=0.115\textwidth]{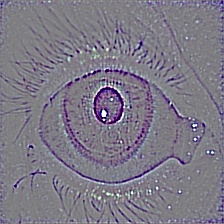}\hskip1mm
	\includegraphics[width=0.115\textwidth]{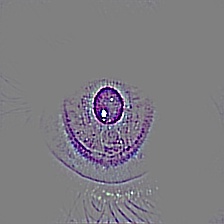}
	\caption{Same as in Fig. \ref{fig:bias-analysis-cam-cold}, but this \textbf{post-mortem sample was misclassified as a live one}.} 
	\label{fig:bias-analysis-cam-cold-failure}
\end{figure}

\begin{figure}[t!]
	\centering
	\includegraphics[width=0.115\textwidth]{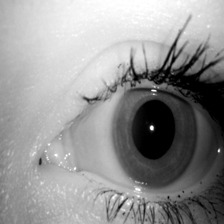}\hskip1mm
	\includegraphics[width=0.115\textwidth]{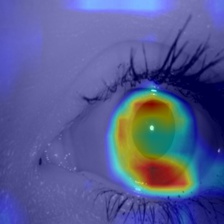}\hskip1mm
	\includegraphics[width=0.115\textwidth]{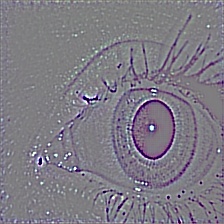}\hskip1mm
	\includegraphics[width=0.115\textwidth]{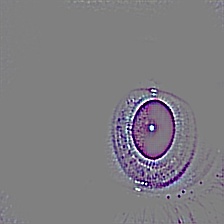}\\\vskip1mm
	\includegraphics[width=0.115\textwidth]{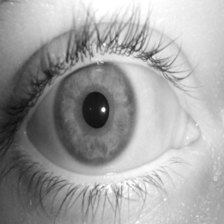}\hskip1mm
	\includegraphics[width=0.115\textwidth]{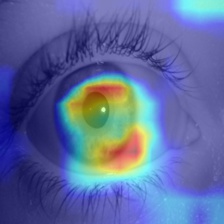}\hskip1mm
	\includegraphics[width=0.115\textwidth]{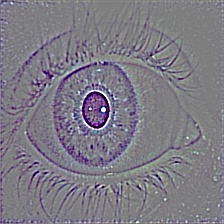}\hskip1mm
	\includegraphics[width=0.115\textwidth]{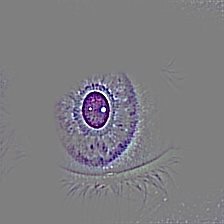}
	\caption{Same as in Fig. \ref{fig:bias-analysis-cam-cold}, but for \textbf{samples of live irises}.} 
	\label{fig:bias-analysis-cam-warm}
\end{figure}

Example predictions were then obtained for both the \emph{live} and \emph{post-mortem} classes, Figs. \ref{fig:bias-analysis-cam-cold} and \ref{fig:bias-analysis-cam-warm}. This shows that the metal parts of the medical equipment used to open the eyelids of deceased subjects indeed {\bf provide class discriminatory cues} (cf. top row in Fig. \ref{fig:bias-analysis-cam-cold}), which in this case is strongly undesirable, as we want our network to recognize post-mortem irises, and not the equipment that may, or may not accompany the post-mortem data collection. Also, for the second investigated post-mortem sample, for which no metal parts are visible, but instead a heavily distorted eyelids are present, the network also pays attention not to the iris itself, but rather to its surrounding -- in this case, the eyelids, cf. bottom row in Fig. \ref{fig:bias-analysis-cam-cold}. In none of these cases can we see strong activations by the iris region. However, when distinctive features such as metal retractors or heavily distorted eyelids are absent from the image, and its overall resemblance to an image of a live iris is strong, the model fails to make a correct prediction, and focuses on the iris region instead, Fig. \ref{fig:bias-analysis-cam-cold-failure}.

Notably, a similar behavior can be observed for images of live irises. When analyzing example activation maps for the \emph{live} class, we see that it is the iris region that produces the strongest activations, while the iris surroundings remain mostly unused, as depicted in Fig. \ref{fig:bias-analysis-cam-warm}. This behavior, contrary to what we observe with post-mortem samples, can be considered desirable.

\subsection{Dataset modification to counteract the bias}

To force-shift the network's attention to the iris, and not its neighborhood, we have manually segmented all images in both datasets, approximating the outer iris boundary with a circle with a radius $R_i$ and then cropping and masking the image to the size of $1.2 R_i$, see Fig. \ref{fig:samplesFromDataset}. This margin of $0.2 R_i$ is preserved purposefully, to represent the differences in the iris surroundings and in the iris-sclera boundary, and not only these present in the tissue itself. The same reasoning is behind leaving the pupillary region unmasked, as the appearance of the pupil can bear liveness information as well. The raw images obtained from the sensor are thus referred to as the \emph{original dataset}, while the modified version is called the \emph{cropped\_masked dataset}. 

\begin{figure}[t!]
	\centering
	\includegraphics[width=0.115\textwidth]{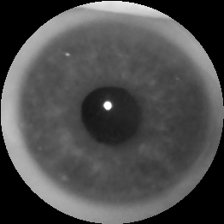}\hskip1mm
	\includegraphics[width=0.115\textwidth]{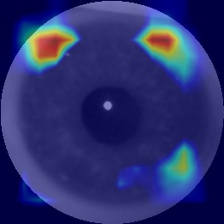}\hskip1mm
	\includegraphics[width=0.115\textwidth]{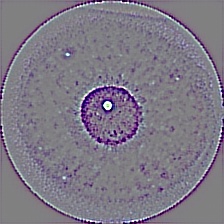}\hskip1mm
	\includegraphics[width=0.115\textwidth]{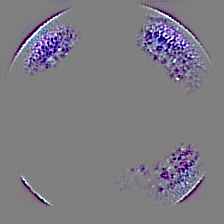}\\\vskip1mm
	\includegraphics[width=0.115\textwidth]{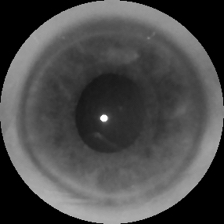}\hskip1mm
	\includegraphics[width=0.115\textwidth]{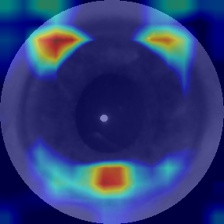}\hskip1mm
	\includegraphics[width=0.115\textwidth]{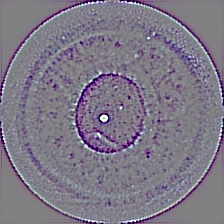}\hskip1mm
	\includegraphics[width=0.115\textwidth]{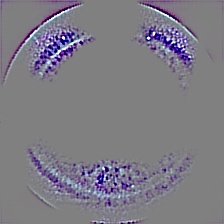}
	\caption{Example class activation maps obtained using the Grad-CAM technique for samples from the \emph{cropped\_masked} dataset, with a model trained on the \emph{cropped\_masked} dataset as well. From left to right: (1) original image, (2) Grad-CAM, (3) guided back-propagation, (4) a combination of (2) and (3). \textbf{Post-mortem samples are represented here, both correctly classified.}} 
	\label{fig:bias-analysis-cam-cold-cropped}
\end{figure}

\begin{figure}[t!]
	\centering
	\includegraphics[width=0.115\textwidth]{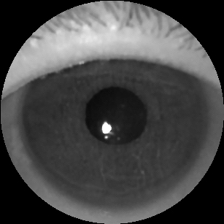}\hskip1mm
	\includegraphics[width=0.115\textwidth]{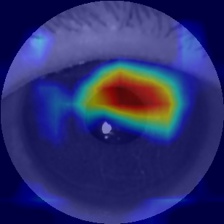}\hskip1mm
	\includegraphics[width=0.115\textwidth]{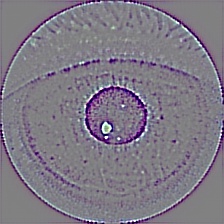}\hskip1mm
	\includegraphics[width=0.115\textwidth]{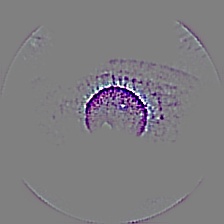}
	\caption{Cropped version of the post-mortem sample misclassified earlier in Fig. \ref{fig:bias-analysis-cam-cold-failure}. The classification is now \textbf{correct}.} 
	\label{fig:bias-analysis-cam-cold-failure-cropped}
\end{figure}

\begin{figure}[t!]
	\centering
	\includegraphics[width=0.115\textwidth]{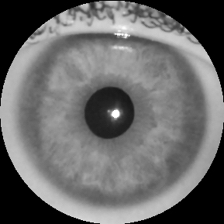}\hskip1mm
	\includegraphics[width=0.115\textwidth]{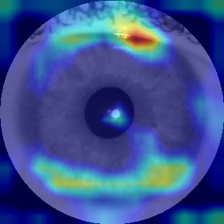}\hskip1mm
	\includegraphics[width=0.115\textwidth]{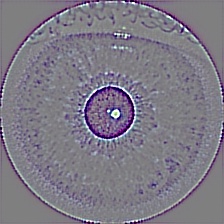}\hskip1mm
	\includegraphics[width=0.115\textwidth]{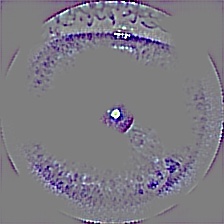}\\\vskip1mm
	\includegraphics[width=0.115\textwidth]{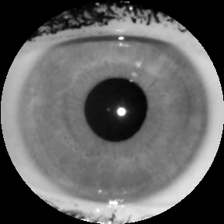}\hskip1mm
	\includegraphics[width=0.115\textwidth]{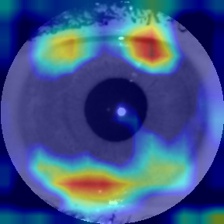}\hskip1mm
	\includegraphics[width=0.115\textwidth]{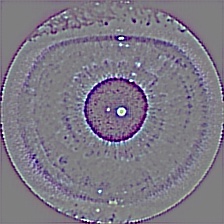}\hskip1mm
	\includegraphics[width=0.115\textwidth]{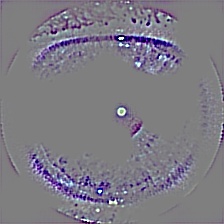}	\caption{Same as in Fig. \ref{fig:bias-analysis-cam-cold-cropped}, but for \textbf{samples of live irises}.} 
	\label{fig:bias-analysis-cam-warm-cropped}
\end{figure}

To validate this reasoning, we train the same model that was employed for assessing class activation maps in the unmodified data, but with the \emph{cropped\_masked} images instead. Activation maps drawn for example \emph{cropped\_masked} iris images are shown in Figs. \ref{fig:bias-analysis-cam-cold-cropped}, \ref{fig:bias-analysis-cam-cold-failure-cropped}, and \ref{fig:bias-analysis-cam-warm-cropped}. As for the \emph{post-mortem} samples, the new model now mostly focuses on the iris and its boundary, cf. Fig. \ref{fig:bias-analysis-cam-cold-cropped}, which seems reasonable, as the iris-sclera boundary is quickly getting blurry as time since death progresses. This is different for the problematic sample examined earlier in Fig. \ref{fig:bias-analysis-cam-cold-failure}, for which the activation map is centered near the pupillary region of the eye. This sample, however, is now correctly classified as a post-mortem one. 

When it comes to samples representing live eyes, the network also seems to assess their PAD score by analyzing the iris-boundary region, but to some degree it also brings its attention to the iris itself, and to the specular reflection found in the middle of the pupil, as depicted in Fig. \ref{fig:bias-analysis-cam-warm-cropped}. These features seem to offer enough discriminatory power to successfully differentiate between the post-mortem and live samples, as the validation accuracy on the \emph{cropped\_masked} subset of subject-disjoint iris images reaches 100\%, compared to less than 95\% obtained for the unmodified set of the same iris images. This also shows that we have managed to successfully shift the attention of the network towards the iris features and discriminative information they offer in recognizing cadaver samples.

\section{Proposed methodology and evaluation}
\label{sec:experiments}

\subsection{Model architecture}
For our solution, we employed the VGG-16 model pre-trained on natural images from the ImageNet database \cite{VGGSimonyanCNNsForRecognition2014}, which has been shown to repeatedly achieve excellent results in various classification tasks after minor adaptation and re-training. We thus performed a simple modification to the last three layers of the original graph to reflect the nature of our binary classification into \emph{live} and \emph{post-mortem} types of images, and performed transfer learning by fine-tuning the network weights to our dataset of iris images representing both classes. 

\subsection{Training and evaluation procedure}
For the network training and testing procedure, 20 subject-disjoint train/test data splits were created by randomly assigning the data from 3 subjects to the test subset, and the data from the remaining subjects to the train subset, both for the \emph{live} and \emph{post-mortem} parts of the database. These twenty splits were made with replacement, making them statistically independent. The network was then trained with each train subset independently for each split, and evaluated on the corresponding test subset. This procedure gives 20 statistically independent evaluations and allows to assess the variance of the estimated error rates. The training, encompassing 10 epochs in each of the train/test split run, was performed with stochastic gradient descent as the minimization method with momentum $m=0.9$ and learning rate of 0.0001, with the data being passed through the network in mini batches of 16 images.

During testing, a prediction of the \emph{live} or \emph{post-mortem} class-wise probability was obtained from the Softmax layer, together with a corresponding predicted categorical label. The probabilities of \emph{post-mortem} samples belonging to their class are also associated with a time that elapsed since death until the sample acquisition. This allow the analysis of classification accuracy in respect to the \emph{post-mortem} time horizon. 

\begin{figure}[t]
	\centering
	\includegraphics[width=0.42\textwidth]{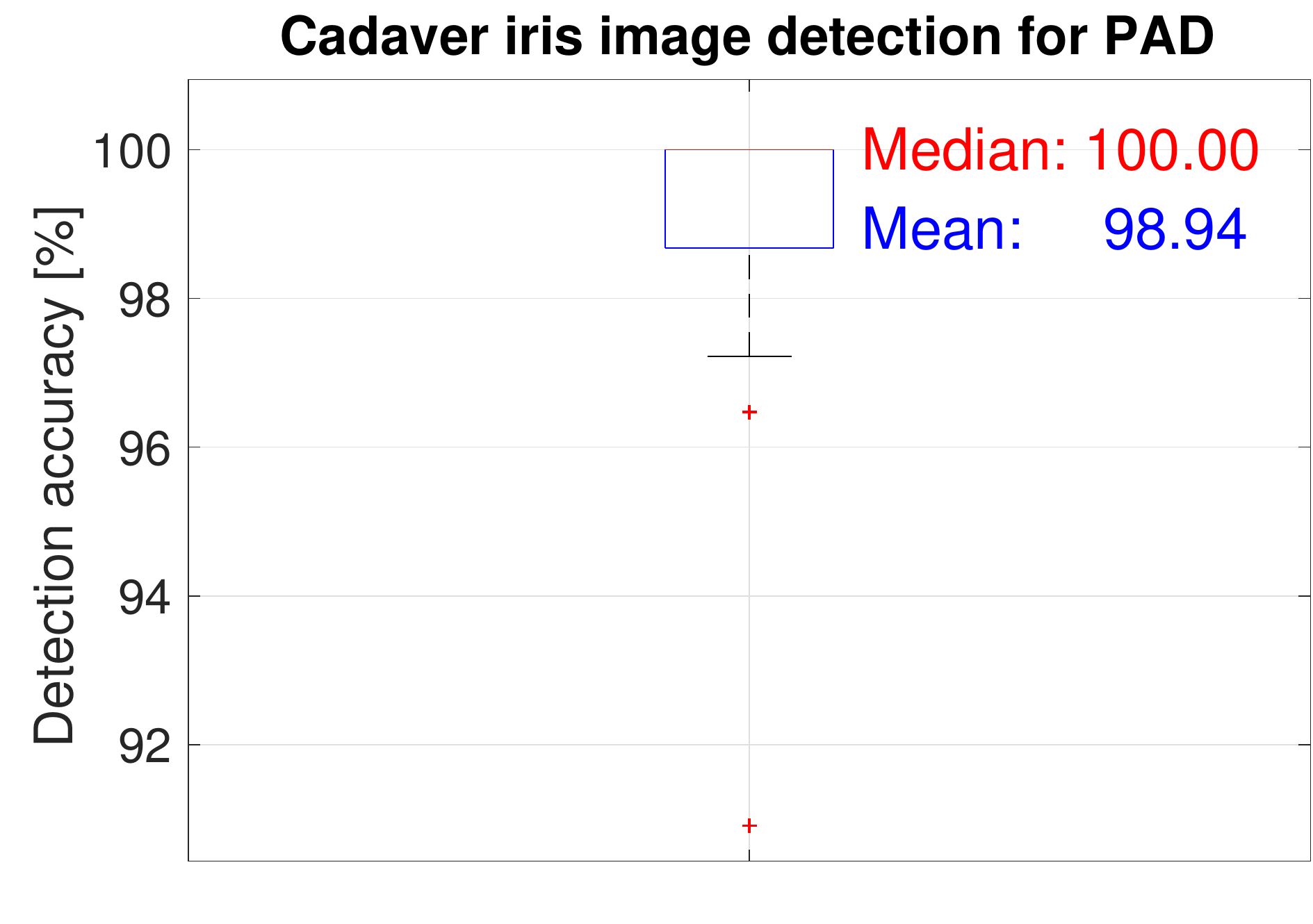}\vskip-4mm
		\caption{Accuracy of classification into live and post-mortem classes, achieved in 20 independent, subject-disjoint train/test data splits.}
	\label{fig:accuracy-averaged}
\end{figure}

\section{Results}
\label{sec:results}

\begin{figure*}[t]
	\centering
	\includegraphics[width=0.99\textwidth]{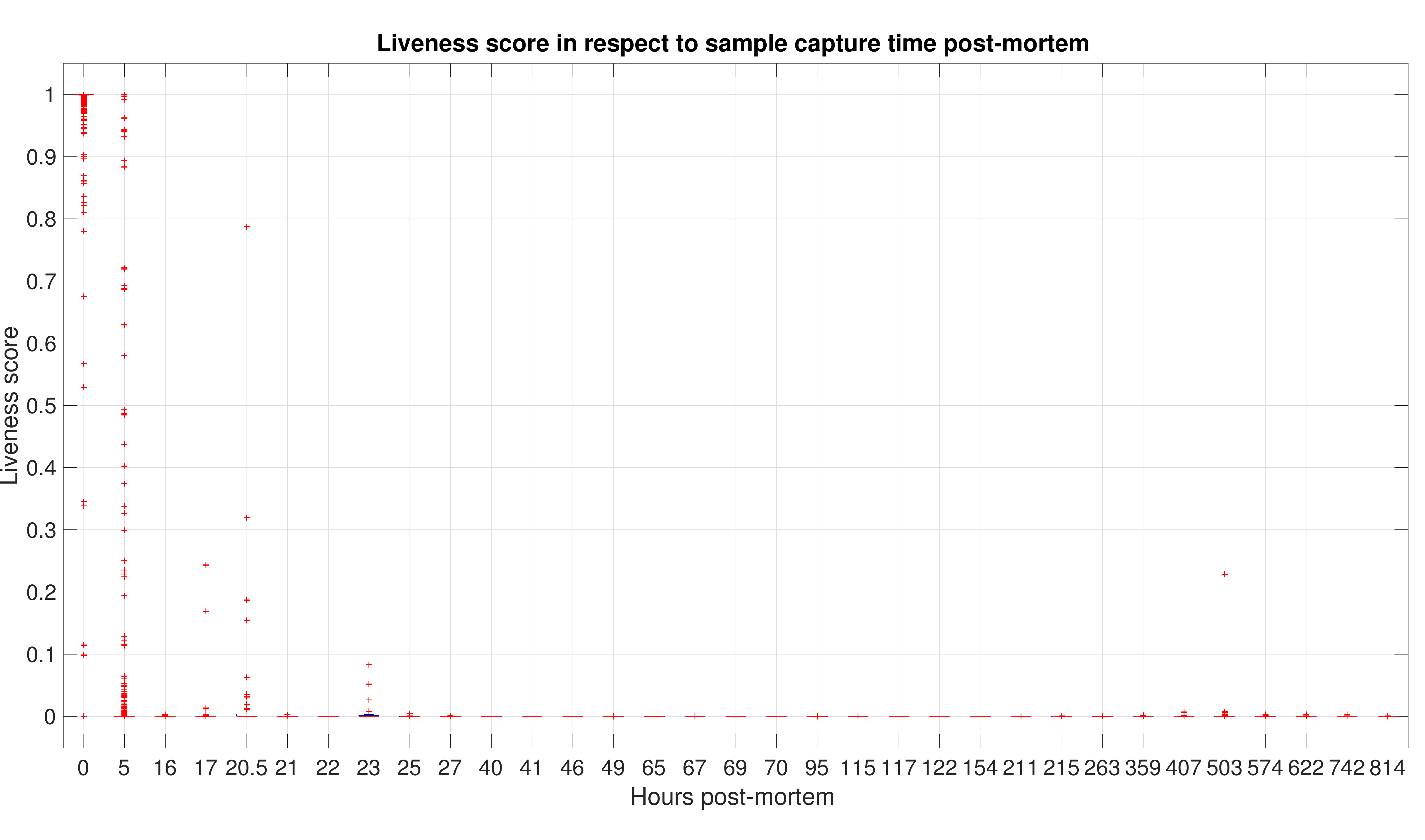}\vskip-5mm
		\caption{Boxplots representing differences in liveness scores earned by the samples acquired in different moments after death. {\bf The lower the score, the more likely the sample represents a post-mortem eye}. Samples denoted as acquired \textbf{zero hours} post-mortem are those collected from live irises.}
	\label{fig:accuracy-timewise}
\end{figure*}

\begin{figure}[t]
	\centering
	\includegraphics[width=0.42\textwidth]{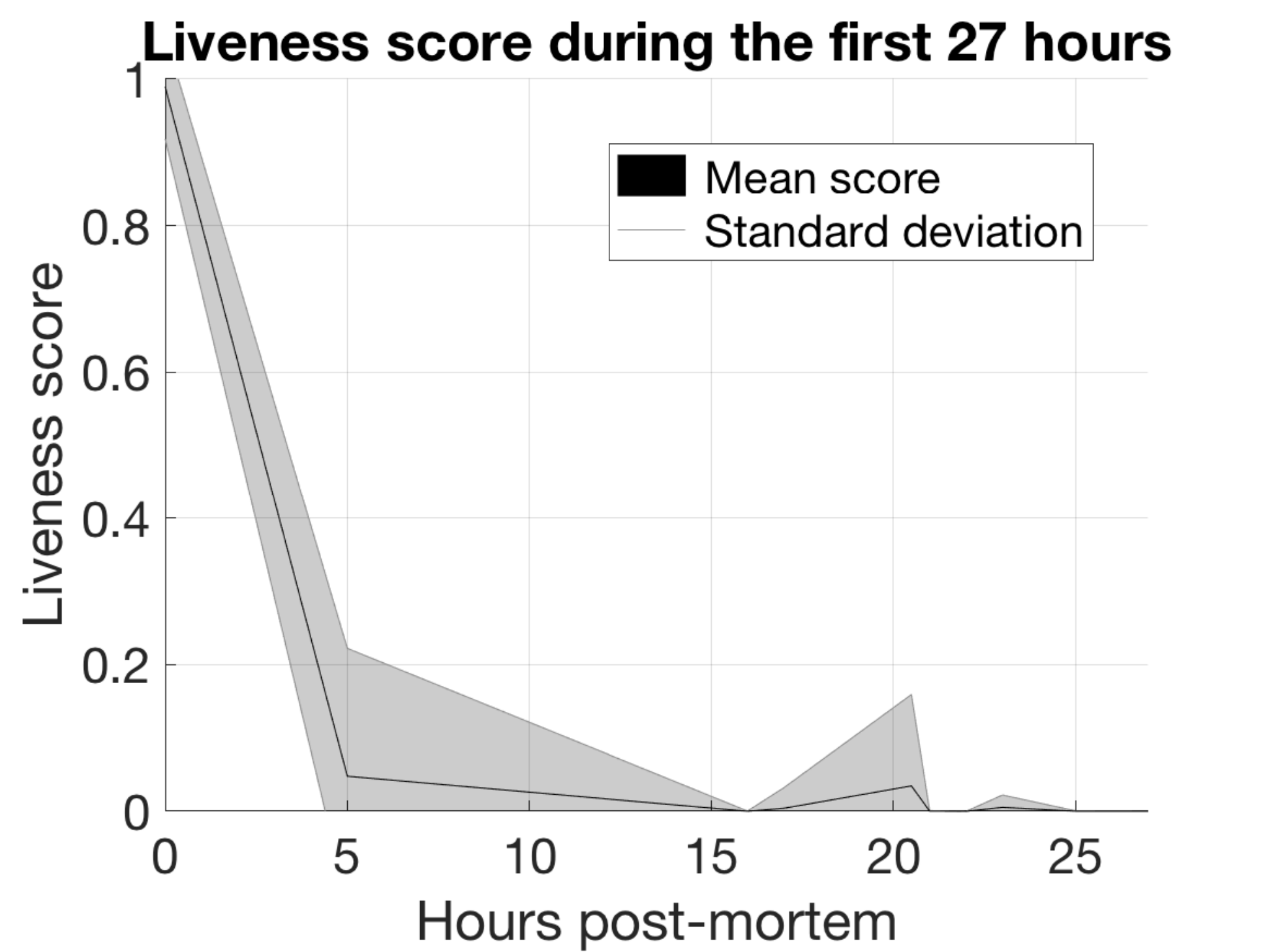}\vskip-2mm
		\caption{Means (solid black line) and standard deviations (grey areas) of liveness score predicted by the network for samples obtained up to 27 hours after death. Samples denoted as acquired \textbf{zero hours} post-mortem are those collected from live irises.} 
	\label{fig:accuracy-timewise-short}
\end{figure}

\begin{figure}[t]
	\centering
	\includegraphics[width=0.42\textwidth]{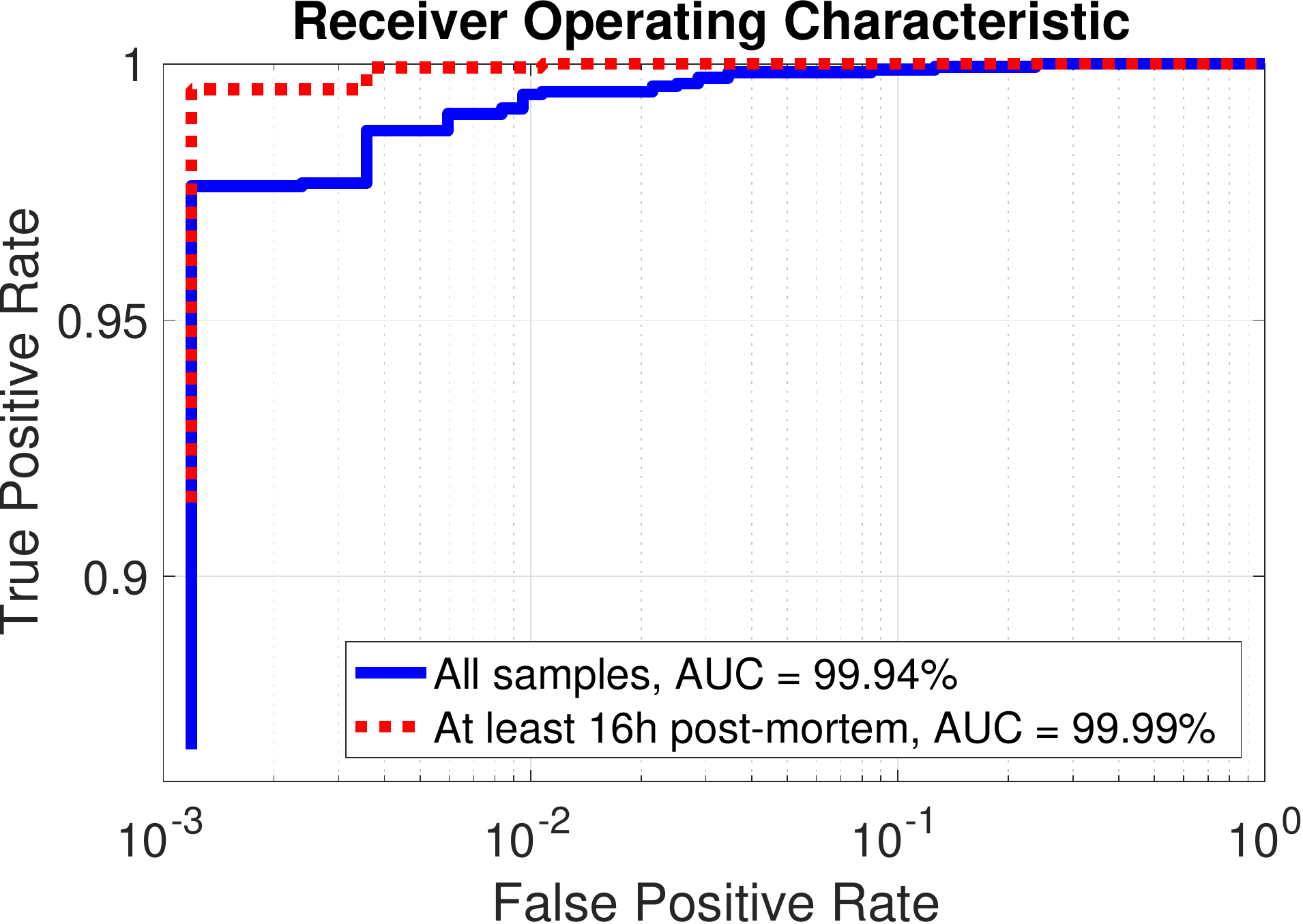}\vskip-1mm
		\caption{ROC curve showing classification accuracy averaged over 20 independent, subject-disjoint train/test data splits. AUC denotes the Area Under Curve metric.}
	\label{fig:ROC-averaged}
\end{figure}

\subsection{Averaged classification accuracy}

Several metrics can be utilized to evaluate the classification accuracy of our solution. First, we average the accuracy achieved in each of the twenty train/test splits, measured as a share of correctly assigned labels to the overall number of trials in a given split, Fig. \ref{fig:accuracy-averaged}. Notably, in most of the splits, the solution achieves a 100\% classification accuracy on the test subset, with the average of 98.94\%.

\subsection{Global performance by Receiver Operating Characteristic}
Receiver Operating Characteristic curves (ROC) are often employed for visualizing performance of classification systems, especially when binary classification is in place. Thus, we present ROC graphs with Areas Under Curve (AUC) calculated for each curve in Fig. \ref{fig:ROC-averaged}. The obtained classification accuracy is almost perfect on the whole set of samples (AUC=99.94\%), and even better when samples acquired shortly (\ie five hours) after death are excluded from the dataset (AUC=99.99\%). We elaborate more on this exclusion in the following paragraph.

\subsection{Classification accuracy in respect to post-mortem time horizon}

Having post-mortem samples acquired for different subjects at multiple time points after death, makes it possible to analyze the performance of our solution in respect to the time that has passed since a subject's demise, Fig. \ref{fig:accuracy-timewise}. Interestingly, and perhaps accordingly to a common sense, the probability of a sample being post-mortem increases as time since death elapses. We can expect a few false matches (post-mortem samples being classified as live iris samples) with images obtained 5 hours after death, regardless of the chosen threshold. This can be attributed to the fact that these images are very similar to those obtained from live individuals, as post-mortem changes to the eye are still not pronounced enough to allow for a perfect classification accuracy. Score means and standard deviations in a close-up of the time period occurring shortly after death is shown in Fig. \ref{fig:accuracy-timewise-short}, which shows that the increased variations in the score distributions in several time points, with the largest variance occurring 5 hours after death. However, the already good accuracy is getting close-to-perfect when these samples are not taken into consideration, Fig. \ref{fig:accuracy-averaged}, red dotted line. Attack Presentation Classification Error Rate (APCER, misclassifying post-mortem samples as live ones) equal to zero can be achieved in such scenario, provided that an appropriate acceptance threshold is established. Bona Fide Presentation Classification Error Rate (BPCER, misclassifying live samples as post-mortem ones) in such case is only about 1\%.

\section{Conclusions}
\label{sec:conclusions}
This paper offers the first, know to us, method for iris liveness detection in respect to the post-mortem setting, based on a deep convolutional neural network VGG-16, adapted and fine-tuned to the task of discerning live and dead irises. The proposed method is able to correctly classify nearly 99\% of the samples, assigning \emph{alive} or \emph{post-mortem} labels, respectively.

Another interesting insight is that samples collected briefly after death (\ie five hours in our study) can fail to provide post-mortem changes that are pronounced enough to serve as cues for liveness detection. However, when classifying samples collected at least 16 hours after death, we found that employing a well-tuned threshold enables APCER=0\%, meaning that no post-mortem sample gets mistakenly classified as a live one, with a probability of misclassifying a live sample as a dead one being around BPCER=1\%. This shows that while post-mortem iris images are relatively easy to identify, those obtained very shortly after a subject's demise can pose problems for automatic solutions due to post-mortem changes not being prominent enough yet.

A significant portion of this paper is also dedicated to an attempt to explain the reasoning behind the decisions provided by our DCNN-based solution, especially with respect to the iris regions that are considered when making these decisions. By employing the Grad-CAM class activation mapping technique, we managed to show that the image regions bearing the most useful discriminatory cues for liveness detection are those containing iris-sclera boundary, and to some extent also the pupillary region.            

We follow the guidelines on research reproducibility by providing a) the source codes, b) trained network weights, and c) the complementary dataset of live irises to all interested researchers, who would like to reproduce the results presented in this paper. We hope that this contribution will stimulate further research in iris presentation attack detection.

\section*{Acknowledgment}
The authors would like to thank Kasia Roszczewska for allowing us to execute the calculations on her GPU, as well as NVIDIA for providing a GPU unit for our laboratory.

\end{document}